%% file: 4dfluid_review.tex
\documentclass[10pt,twocolumn,letterpaper]{article}

\usepackage{iccv}
\usepackage{times}
\usepackage{epsfig}
\usepackage{graphicx}
\usepackage{amsmath}
\usepackage{amssymb}
\usepackage{xcolor}
\usepackage{bm}
\usepackage{float}
\usepackage{overpic}
\usepackage{caption}

\usepackage[pagebackref=true,breaklinks=true,letterpaper=true,colorlinks,bookmarks=false]{hyperref}
\usepackage{cleveref}
\Crefname{equation}{Eq.}{Eqs.}
\Crefname{section}{Sec.}{Sec.}

\iccvfinalcopy 

\newcommand{\neural}[2]{#1_{\bm{#2}}}
\newcommand{\real}{\mathbb{R}}
\newcommand{\expnumber}[2]{{#1}\mathrm{e}{#2}}
\newcommand{\loss}[1]{\mathcal{L}_{\text{#1}}}

\DeclareMathOperator*{\argmin}{argmin}

\ificcvfinal\fi

\begin{document}

\makeatletter
\let\@oldmaketitle\@maketitle
\renewcommand{\@maketitle}{\@oldmaketitle
  \input{imgs/new_teaser/figure}\bigskip}
\makeatother

\title{Fluid Dynamics Network: \\ Topology-Agnostic 4D Reconstruction via Fluid Dynamics Priors}

\author{Daniele Baieri\\
Sapienza University of Rome\\
{\tt\small baieri@di.uniroma1.it}
\and
Stefano Esposito\\
Hochschule Bonn-Rhein-Sieg\\
{\tt\small stefano.esposito@h-brs.de}
\and
Filippo Maggioli\\
Sapienza University of Rome\\
{\tt\small maggioli@di.uniroma1.it}
\and
Emanuele Rodol\`a\\
Sapienza University of Rome\\
{\tt\small rodola@di.uniroma1.it}
}

\maketitle
\ificcvfinal\thispagestyle{empty}\fi

\begin{abstract}
    Representing 3D surfaces as level sets of continuous functions over $\real^3$ is the common denominator of neural implicit representations, which recently enabled remarkable progress in geometric deep learning and computer vision tasks. 
    In order to represent 3D motion within this framework, it is often assumed (either explicitly or implicitly) that the transformations which a surface may undergo are homeomorphic: this is not necessarily true, for instance, in the case of fluid dynamics. 
    In order to represent more general classes of deformations, we propose to apply this theoretical framework as regularizers for the optimization of simple 4D implicit functions (such as signed distance fields). We show that our representation is capable of capturing both homeomorphic and topology-changing deformations, while also defining correspondences over the continuously-reconstructed surfaces.
    
   
\end{abstract}

\section{Introduction}

Representing the motion of 3D objects is a crucial aspect in several areas of 3D graphics. For instance, skinning models are an established method for mesh animation, which provides a plausible and efficient way of deforming a surface given a sequence of skeleton poses~\cite{jacobson:2014:skinning}. However, various applications such as fluid simulation require representing topology-changing (formally, non-homeomorphic) deformations, which are notoriously difficult to approach with mesh-based approaches.


Implicit representations are becoming an increasingly important topic in computer graphics, as they enable or improve a variety of game-changing technologies, such as shape modeling~\cite{malladi:1995:sdfmodeling}, accurate shape and scene reconstruction~\cite{icml2020_2086,mildenhall2020nerf} and fast rendering~\cite{esposito:2022:kiloneus,hart1993sphere,mueller2022instant}. Yet, the representation of a 3D object with a signed distance field introduces many challenges since many established pipelines from classical computer graphics cannot be applied anymore. Between these challenges, the non-rigid animation of a body is one of the most obvious, as standard approaches like skinning models do not have straightforward conversions for working with implicit representations.

\begin{figure}[h]
\input{imgs/torus-water/figure}
\end{figure}

Recently, some works presented significant progresses in this direction. Taking advantage of deep learning techniques, it has been shown that implicit surfaces can be successfully rigged and animated using skeletons, with a particular focus on humanoid characters~\cite{chen2021snarf,tiwari21neuralgif}. Other approaches got rid of skeletal animations and focused on reconstructing 3D geometries that are continuously deformed over time from a set of discrete observations~\cite{Lei2022CaDeX,OccupancyFlow}. However, these techniques represent deformation as continuous bijective functions with continuous inverse, and as a consequence they are limited to representing fixed-topology deformations.

We fill this gap by introducing a new computational model that can reconstruct continuously deformable 3D objects from a discrete sample of observations without any assumptions on the topology changes. Our approach is shown to be effective in settings where the geometry deformation is homeomorphic (such as character animations), as well as applications where the topology changes drastically over time (like fluid dynamics scenes). We also show that the 4D reconstruction produced by our model can be used to track dense correspondences in the evolving surface, making it also suitable for shape-matching applications.


\section{Related work}


\paragraph{Neural implicit representations}

Representing 3D surfaces as continuous functions over ambient space is a recent trend in geometric deep learning. Manipulating continuous surfaces offers new perspectives in geometry processing, and led to significant advances in various complex tasks. Occupancy fields \cite{Occupancy_Networks} and signed distance fields \cite{Park_2019_CVPR} were applied in several different settings, such as 3D surface reconstruction \cite{Atzmon_2020_CVPR,icml2020_2086,Peng2020ECCV,sitzmann2019siren,wang2021neus}, animation \cite{chen2021snarf,tiwari21neuralgif}, and inverse rendering \cite{DVR,oechsle2021unisurf,yariv2020multiview}. This last application was revolutionized by neural radiance fields \cite{mildenhall2020nerf}, which inspired an entire line of research. Successively, NeRFs were applied to deforming scenes (4D reconstruction from multi-view videos) \cite{chu2022physics,park2021nerfies,dnerf,tretschk2020nonrigid}, MRI scans \cite{coronafigueroa2022mednerf,nerfmrisupersample}, and for scene relighting \cite{boss2021nerd,nerv2021,zhang2021nerfactor}. Instant-NGP \cite{mueller2022instant} further enabled application of neural implicit representations by showing how to optimize them in a matter of seconds.

\begin{figure}[h]
    \input{imgs/fluid-dfaust-examples/figure}
\end{figure}


\paragraph{4D shape reconstruction}

Continuous 4D reconstruction methods tackle the reconstruction of dynamic scenes both in space and time from a sequence of sparse observations, typically point-clouds or 2D images. While 3D reconstruction allows for several algorithmic baselines (such as \cite{ssdrecon,poisson}), the complexity introduced by the requirement of reconstructing motion steered this task towards deep learning methods, and most recently, neural implicit representations.

PSGN \cite{psgn} is a 3D reconstruction method for multi-view images which can be easily generalized to predict a 4D point cloud, i.e. the point cloud trajectory instead of a single point set. Similarly, implicit representation networks such as occupancy net \cite{Occupancy_Networks} can be generalized by adding an input coordinate (usually called ONet4D), in order to define the occupancy field in the spatio-temporal domain by predicting time-varying occupancy values. By assigning a velocity vector to every point in space and time, Niemeyer \etal \cite{OccupancyFlow} offered a method to represent continuous 4D shapes by trasporting a static occupancy field, while also implicitly modeling dense correspondences. Tang \etal \cite{tang2021learning} proposed a novel 4D point cloud encoder design that performs efficient spatio-temporal shape properties aggregation from 4D point cloud sequences, improving upon Occupancy Flow's reconstructed geometry. CaDeX \cite{Lei2022CaDeX} explicitly represents the deformation as an homeomorphism, by factorizing it into a continuous invertible function and its continuous inverse, mapping the source geometry frame into a common 3D coordinate space and back to the destination frame, respectively; its prior allows for topology preservation by construction and the recovery of consistent correspondences across frames.

All aforementioned methods are data driven, where at test time reconstructions are conducted from an image or a partial observations leveraging prior knowledge learned from large-scale datasets containing both complete surfaces (\ie meshes) and their sparse observations. This improves the accuracy and consistency of the results, especially in situations where the input data is noisy or incomplete. However, prior-based methods can be computationally expensive and their large amounts of training data requirement can limit their practical applicability. 

Furthermore, a common assumption among these methods is preservation of topology; for instance, the integration of points through the velocity field learned by Occupancy Flow is also implicitly an homeomorphism. Our method will make no such assumption, thus retaining an additional level of generality, and can be optimized from a sequence of dense oriented point clouds, without data priors.






\begin{figure}[h]
\input{imgs/cold-gas-obstacles/figure}
\end{figure}

\paragraph{Computational fluid dynamics}

The simulation of fluid dynamics is a fundamental topic in computer graphics, and much research has been devoted to this area in the last decades. Researchers have approached this problem from a variety of points of view, spanning from the direct simulation of Navier-Stokes equations~\cite{bridson:2007:fluidsiggraph} to non physics-based algorithm for credible visual results~\cite{fournier:1986:oceanwaves,hisinger:2002:oceanwaves}. For a detailed survey on fluid solvers, we refer to Bridson's textbook~\cite{bridson:2018:fluid} and Koshier \emph{et al.}~\cite{koshier:2019:fluid}. Despite the amount of work dedicated to this problem, the room for improvement is still significant, and new studies on fluid simulation are continuously coming out. In recent years, many studies tried to achieve better quality in simulation by focusing on specific problems, like certain kinds of waves~\cite{Huang:2021:VastOcean} and particular environmental conditions~\cite{lan:2022:smokesubway}. Simultaneously, new techniques for speeding up simulations have been developed, with particular attention to the interaction between solid obstacles and turbulent fluids~\cite{li:2020:turbulentflow,lyu:2021:fluidsolid} and to efficient space partitioning~\cite{Shao:2022:Multigrid}.

\paragraph{Machine learning and fluid simulation}
Recently, more innovative approaches tried to combine the descriptive power of machine learning with the simulation of fluid dynamics. Um \etal showed a general method to improve PDE solvers (such as fluid simulators) solutions using statistical ML models \cite{solverintheloop}. Vinuesa \emph{et al.}~\cite{vinuesa:2022:fluidlearning} tried to apply learning paradigms to different stages of fluid simulation, showing the potential improvement in quality and performance. Following this direction, data-driven models have been developed for upsampling turbulent flows and obtaining fine-grained details from coarse simulations~\cite{bai:2020:upsamplingsmoke,bai:2021:predictingturbulence}. Fluid super-resolution has also been tackled in a GAN paradigm, as in \cite{fluidsuperresgan,xie2018tempoGAN}. A completely different approach has been proposed in Li \emph{et al.}~\cite{li:2022:neuralnetfluid}, where particle-based fluid models have successfully been related to graph neural networks obtaining performance improvement at basically no quality cost. Ummenhofer \etal \cite{Ummenhofer2020Lagrangian} later showed that the same can be achieved without graph structure, by performing continuous convolution over sets of points. More similar in spirit to our work is Chu \emph{et al.}~\cite{chu2022physics}: the authors apply fluid simulation priors to inverse rendering of dynamic fluid scenes from sparse multi-view videos, without geometric information.


\begin{figure}[h]
\input{imgs/dfaust-reconstruction/figure}
\end{figure}

\section{Method}

\subsection{Preliminaries}

Simulating fluid dynamics typically involves computing a velocity field $\bm{u}$, by integrating the Navier-Stokes PDE:
\begin{align}
    \dfrac{\partial \bm{u}}{\partial t} + (\bm{u}\cdot\nabla)\bm{u} - \nu\nabla^2\bm{u} &= -\dfrac{1}{\rho}\nabla p +\bm{g}\label{eq:navier} \\
    \nabla\cdot\bm{u} &= 0 \label{eq:div}
\end{align}
And applying the resulting velocity field $\bm{u}$ to transport some type of geometric representation. For our purposes, we represent geometry as a scalar field $f$, which can be transported via the advection equation (assuming \cref{eq:div} holds):
\begin{equation}
    \dfrac{\partial f}{\partial t} + \bm{u}\cdot\nabla f = 0\label{eq:advect}
\end{equation}

Typically, simulators will integrate a simpler form of Eq.~\ref{eq:navier}, only enforcing self-advection:
\begin{equation}
    \dfrac{\partial \bm{u}}{\partial t} + (\bm{u}\cdot\nabla)\bm{u} = 0\label{eq:selfadvect}
\end{equation}
and add on top of this partial result the effects of pressure ($p$), external forces ($\bm{g}$) and viscosity ($\nabla^2\bm{u}$), which are solved for independently. This approximation mitigates the complexity of integrating \Cref{eq:navier} while retaining high precision. In a similar spirit, we use \Cref{eq:selfadvect} as a structural prior and allow our model to estimate the influence of the aforementioned components from the data.

\begin{figure}[h]
\input{imgs/viscous-vortex/figure}
\end{figure}


\subsection{Regularizing 4D reconstruction by CFD priors}

Our method leverages fluid simulation priors as a regularizer for the optimization of a neural representation of the 4D geometry and dynamics. Let $\neural{f}{\theta}\colon\real^4\to\real^d$, $\neural{v}{\phi}\colon\real^4\to\real^3$ be our (temporal) geometry and velocity functions, represented as neural networks. As outlined by Niemeyer \etal in \cite{OccupancyFlow}, a trivial solution to the 4D reconstruction task can be found by optimizing the general loss:
\begin{equation}
    \loss{recon} = \sum_{t\in T} L\left(\neural{f}{\theta}, \mathcal{X}_t; \tau(t)\right)\label{eq:lrecon}
\end{equation}
Where $\{\mathcal{X}_t\}_{t\in T}$ are time-labeled sparse geometry observations, $T$ is the set of available supervised time steps, $\tau$ maps a timestep label to a real time value, and $L$ is an error function comparing the optimized function and the ground truth data. 

This model (DeepSDF4D in the following), while technically correct, does not learn any information about the dynamics underlying the observations; thus, it is similar to applying a 3D reconstruction technique for each frame. As a consequence, no type of correspondence among reconstructed surfaces is established during training. Despite these shortcomings, however, it allows to represent general deformations of 3D geometry, including topological and volume changes. Most proposals in 4D reconstruction literature (such as \cite{Lei2022CaDeX,OccupancyFlow}) rule this possibility out by assuming to only work with isometries (either implicitly or explicitly); we discuss how to improve DeepSDF4D in order to model general deformations (up to volume changes), while establishing correspondences between the reconstructed surfaces.

Similarly to Chu \etal \cite{chu2022physics}, we train $\neural{v}{\phi}$ jointly with $\neural{f}{\theta}$ using the following losses, adapted from \Cref{eq:selfadvect,eq:div,eq:advect}:
\begin{align}
    \loss{div} =& \int_{0}^{\tau}\int_{\real^3} \nabla\cdot\neural{v}{\phi} \,d\bm{x}\,dt \label{eq:ldiv} \\
    \loss{advect} =& \int_{0}^{\tau}\int_{\real^3} \dfrac{\partial \neural{f}{\theta}}{\partial t} + \neural{v}{\phi}\cdot\nabla\neural{f}{\theta}  \,d\bm{x}\,dt\label{eq:ladvect}  \\
    \loss{NS} =& \int_{0}^{\tau}\int_{\real^3} \dfrac{\partial \neural{v}{\phi}}{\partial t} + (\neural{v}{\phi}\cdot\nabla)\neural{v}{\phi} \,d\bm{x}\,dt  \label{eq:lns}
\end{align}
Where $\mathcal{L}_{\text{advect}}$ is only used to train $\neural{v}{\phi}$ (otherwise it is trivially minimized by zeroing $\neural{f}{\theta}$). We discuss how we approximate these integrals in \Cref{sec:implement:training}.

While \Cref{eq:ladvect} is enough to optimize $\neural{v}{\phi}$ with respect to a given set of observations (assuming $\neural{f}{\theta}$ is also trained via \Cref{eq:lrecon}), we still need to regularize our geometry function based on our learned velocity field. An option is to perform random linear warping, as in \cite{chu2022physics,tretschk2020nonrigid}: that is, we query our geometry network for points $\bm{x}$ at a given timestep $t_i$ as 
\begin{equation}
\hat{\neural{f}{\theta}}(\bm{x}, t_i) =  \neural{f}{\theta}(\bm{x} + \delta_t\neural{v}{\phi}(\bm{x}, t_i), t_i + \delta_t)\label{eq:linwarp}
\end{equation}
For randomly sampled $\delta_t$ (further discussion about sampling in \Cref{sec:implement:training}). In combination with \Cref{eq:lrecon}, linear warping allows to transport supervision signals across time through $\neural{v}{\phi}$. Finally, we instantiate $\loss{recon}$ as a SDF reconstruction loss, similar to Park \etal \cite{Park_2019_CVPR}: 
\begin{equation}
    L\left(\neural{f}{\theta}, \mathcal{X}_t; \tau(t)\right) = \int_{\real^3} \left\lVert \neural{f}{\theta}(\bm{x}, \tau(t)) - d(\bm{x}, \mathcal{X}_t)  \right\rVert \,d\bm{x} \label{eq:lsdf}
\end{equation}
Where $\{\mathcal{X}_t\}_{t\in T}$ are oriented point clouds, and $d$ computes the ground-truth signed distance value for $\bm{x}$ by mapping it to the closest point $\bm{p}$ on $\mathcal{X}_t$ and determining its sign (inside/outside) via the winding number of $\bm{p}$ \cite{winding}.

\subsection{Implementation details}\label{sec:implement}

\subsubsection{Model}
\looseness=-1

We implement $\neural{f}{\theta}$ and $\neural{v}{\phi}$ as two Siren \cite{sitzmann2019siren} networks with 5 layers and 256 hidden units. We find that the high-frequency bias induced by this architecture is really beneficial in learning erratic and unpredictable dynamics such as fluid behaviour, allowing to considerably speed up convergence and properly model surface details. \Cref{fig:fluid-siren-vs-prev} highlights this aspect by comparing Siren with a simple MLP with positional encoding \cite{tancik2020fourfeat}. Our networks are trained with the loss function
\begin{equation}
    \mathcal{L} = \loss{recon} + \lambda_1\loss{div} + \lambda_2\loss{advect} + \lambda_3\loss{NS}
\end{equation}
Where we set $\lambda_1=\lambda_2=10, \lambda_3=0.1$.

\begin{figure}[h]
    \input{imgs/ball-siren-compare/figure}
\end{figure}

\subsubsection{Training}\label{sec:implement:training}
\looseness=-1

Our networks are trained for 500 epochs using the Adam optimizer~\cite{adam} with a learning rate of $\expnumber{1}{-4}$. During each training iteration, we need to approximate the integrals in our loss functions (\Cref{eq:lsdf,eq:ldiv,eq:ladvect,eq:lns}). We approximate the time integral via stratified sampling, similarly to how Mildenhall \etal \cite{mildenhall2020nerf} integrate radiance and density to compute light transport on a single ray. Over multiple epochs, this helps to retain continuity of the representation over time (as opposed to, \eg, always integrating over a fixed linear space of time steps). Therefore, we evaluate our loss functions independently for each time step and sum all the loss gradients before updating network parameters. For each timestep, the space integral is approximated by uniformly sampling in a bounding box $D$.

Training occurs in three phases: first, we train $\neural{f}{\theta}$ to near convergence over the input data. Then, based on the partial solution provided by $\neural{f}{\theta}$, we train the velocity network $\neural{v}{\phi}$ by the fluid dynamics priors encoded in \Cref{eq:ladvect,eq:lns,eq:ldiv}. During the last phase, we introduce linear warping with a linear schedule, \ie, the interval from which we uniformly sample $\delta_t$ to compute \Cref{eq:linwarp} grows linearly with training time, up to the entire space between adjacent supervised frames. We find that running each phase for 200 epochs yields satisfying results.

\begin{figure}[H]
    \centering
    \includegraphics[width=\columnwidth]{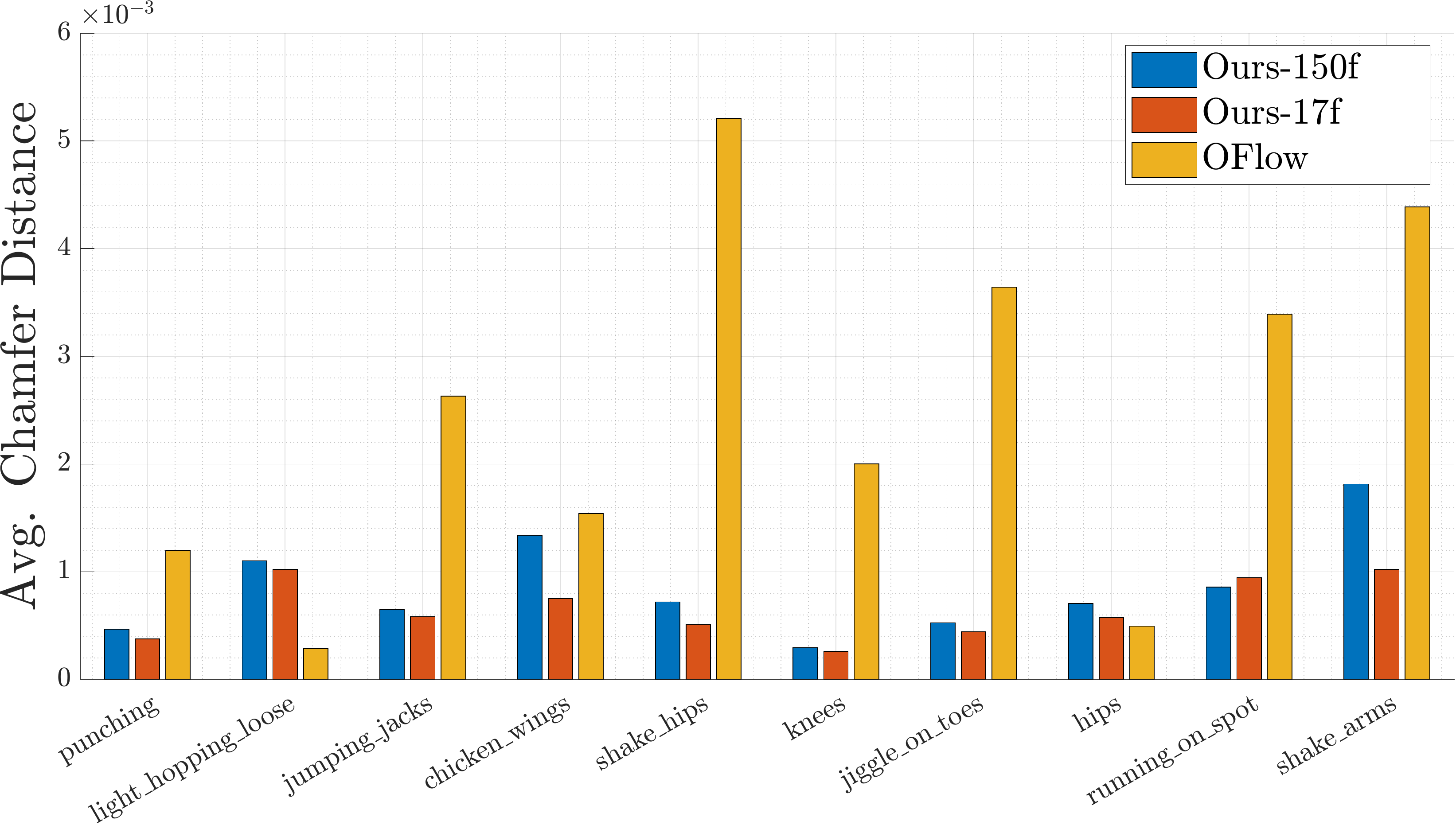}
    \caption{Evaluation of reconstruction quality of our method on a subset of the DFAUST dataset (lower is better). \texttt{OFlow} and \texttt{Ours-17f} are evaluated on the first 17 frames of each sequence. While the high frequency of our geometry function  may hinder reconstruction quality to a small degree, our method manages to reconstruct dynamics more faithfully than Occupancy Flow.}
    \label{fig:recon-quality}
\end{figure}

\section{Experiments}
\looseness=-1

\subsection{4D surface reconstruction}\label{sec:exp1}
\looseness=-1
In order to measure the representational power of our method, we compare meshes extracted from our method using marching cubes \cite{mcubes} with resolution 128 against ground-truth meshes, using the chamfer distance as metric. We run the evaluation over a small subset of the DFAUST dataset \cite{dfaust:CVPR:2017}, where we select one sequence type for each subject, so that most of the sequence types are represented. The results are showed in \Cref{fig:recon-quality}.

\begin{figure}[b]
    \centering
    \includegraphics[width=\columnwidth]{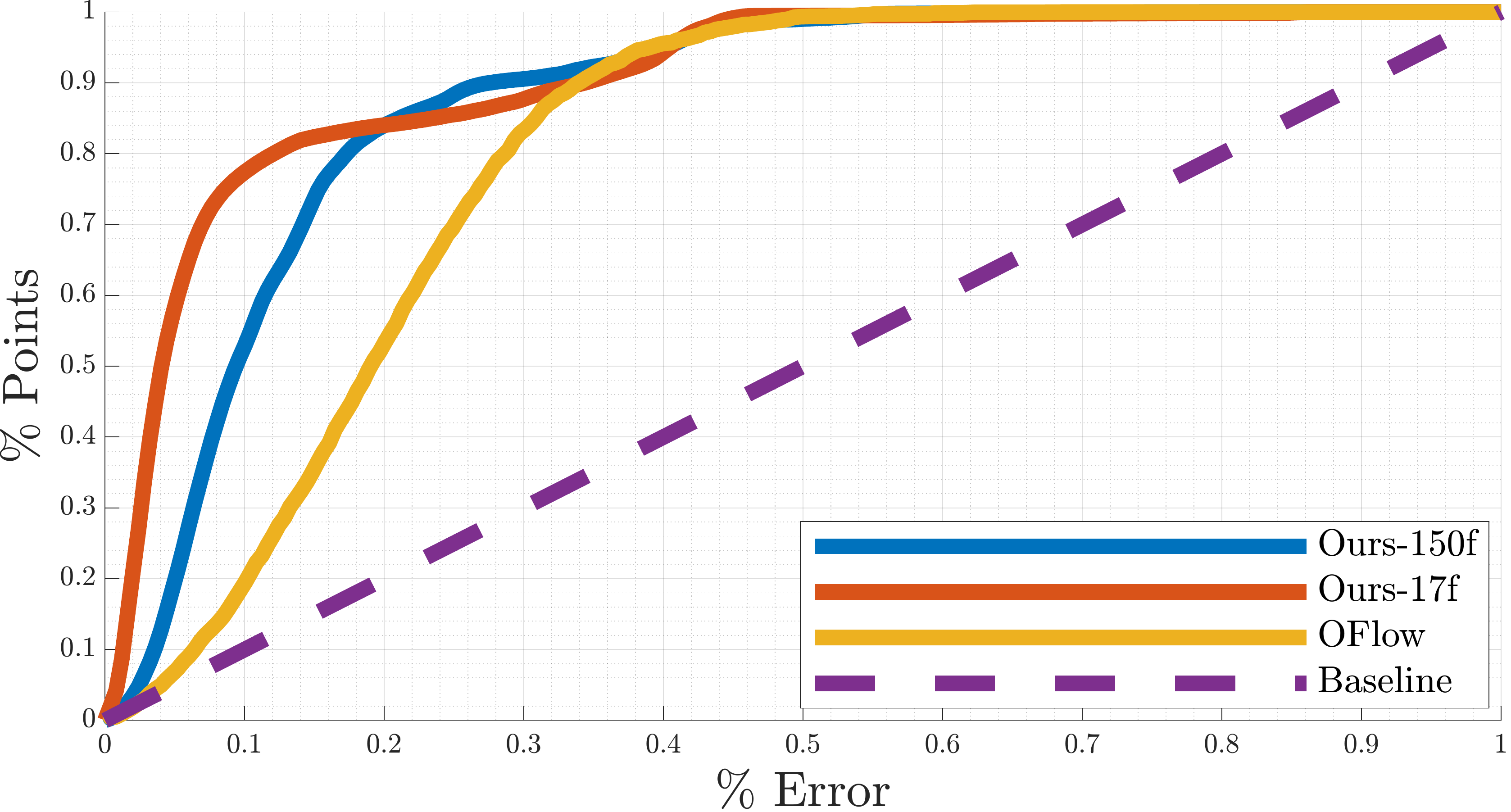}
    \caption{Quantitative comparison of correspondences on DFAUST between our method and Occupancy Flow \cite{OccupancyFlow}. Our method manages to match most points with low error (80\% of points have error percentage below 10\% for the 17 frames evaluation). Saturation seems comparable with the Occupancy Flow pretrained model.}
    \label{fig:compare-dataset-dfaust}
\end{figure}

\subsection{Shape correspondence}

Our method defines a temporal velocity field, which we may leverage in order to define correspondences among the reconstructed surfaces. Furthermore, by relaxing assumptions on the deformation function, we model correspondences with changes of topology (\eg \Cref{fig:non-homeo-matching}) in a fully unsupervised fashion. Formally, for a single sequence of observations $\{\mathcal{X}_i\}_{i=1}^n$, we define the correspondence between the vertices of $\mathcal{X}_1$ and $\mathcal{X}_n$ via the initial value problem:

\begin{align}
\begin{split}
    \Phi(\mathcal{X}_1, 0) &= \mathcal{X}_1 \\
    \dot{\Phi}(\mathcal{X}_1, t) &= \neural{v}{\phi}(\mathcal{X}_1, t)
\end{split}
\end{align}

We integrate $\Phi$ via Euler using $[\tau(1); \tau(n)]$ as time interval. After flowing a point $x$ through the entire time interval via $\Phi(x, \tau(n))$, we select the result on the target set of vertices as the nearest neighbor

\begin{equation}
    \Pi(x) = \argmin_{z\in\mathcal{X}_n} \lVert \Phi(x, \tau(n)) - z \rVert_2
\end{equation}

In order to compare our matching method quantitatively to other 4D reconstruction algorithms with correspondences, we plot a comparison in cumulative matching curves evaluated on the same subset of DFAUST described in \Cref{sec:exp1}. For each analyzed sequence, we match the first and last frames; then, we take as point-wise error metric the geodesic distance between the ground-truth match and the predicted match. The results are showed in \Cref{fig:compare-dataset-dfaust}. We evaluate our method with subsequences of 150 frames and 17 frames, in order to compare it to the pretrained Occupancy Flow model, which was trained only on 17 frames sequences.

\begin{figure}[b]
\input{imgs/non-homeo-matching/figure}
\end{figure}

\section{Conclusions}

\subsection{Discussion and limitations}

We presented a non-data-driven method for 4D reconstruction which is capable of capturing a vast range of geometric deformations and define dense, continuous correspondences among the reconstructed surfaces. 
One main limitation of our method is its requirement for dense input point clouds; we propose to investigate solutions in the future (see \Cref{sec:future}). 

Furthermore, the simplified version of the Navier-Stokes equation which we use in our loss function (\Cref{eq:lns}) is not enough to capture any fluid behaviour: imagine a cylinder rotating on its ``up'' axis. The ground-truth SDF signal for the sequence of shapes will be constant, and therefore the velocity field regressed by our method will be close to zero everywhere. This setting is also not unlikely in fluid simulation, as it is a close analogy for a vortex of high-viscosity fluid. This issue may be addressed by employing the complete Navier-Stokes form, at the cost of having to regress additional functions (such as pressure and external forces) and to optimize second derivatives of the velocity field. One last limitation would be the case of handling non-volume-preserving deformations; however, the entire fluid simulations framework is unsuited for this purpose, thus we consider it to be outside of our scope.

\subsection{Future work}\label{sec:future}

In the near future, we intend to gather the experimental fluid data which we employed for this work into a proper, multimodal (including multi-view renders) 4D geometry dataset. This should provide for a useful benchmark for topology-agnostic methods in geometry processing.

Furthermore, we intend to generalize our presented formulation to a data-driven model: this should allow us to relieve our need for dense geometric input at inference time, albeit constraining the learned model to the distribution presented in the employed dataset.

Lastly, we intend to investigate the capabilities of our formulation in point cloud segmentation, in particular, we believe our method has promise in separating static and dynamic content in 4D point clouds.

{\small
\bibliographystyle{ieee_fullname}
\bibliography{4dfluid_review}
}

\end{document}

%% file: imgs/new_teaser/figure.tex
\centering
\includegraphics[width=\textwidth]{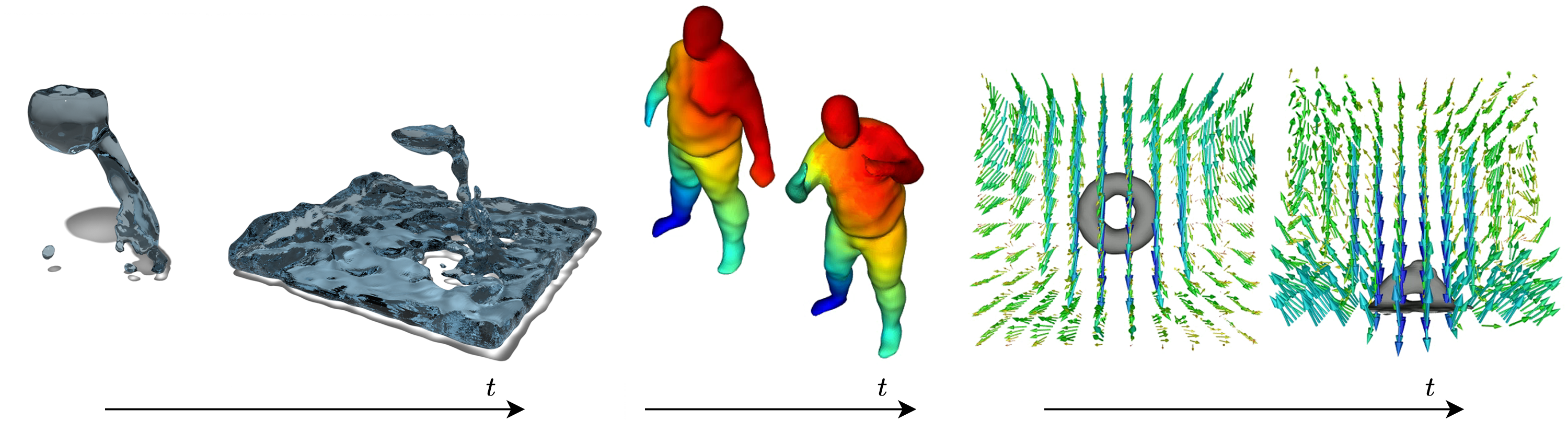}
\captionof{figure}{Left: simulation of a fluid dynamic reconstructed with our method. Middle: a correspondence on a humanoid character induced by our model. Right: a 3D object and the deformation velocity field reconstructed with our method.}
\label{fig:new_teaser}

%% file: imgs/torus-water/figure.tex
\centering
\includegraphics[width=.33\columnwidth]{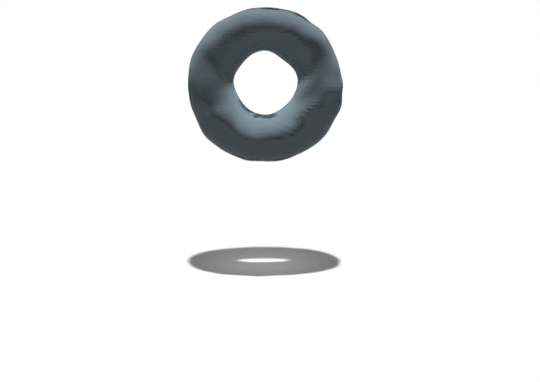}%
\includegraphics[width=.33\columnwidth]{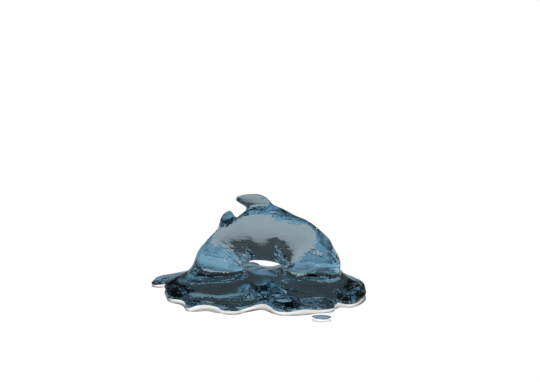}%
\includegraphics[width=.33\columnwidth]{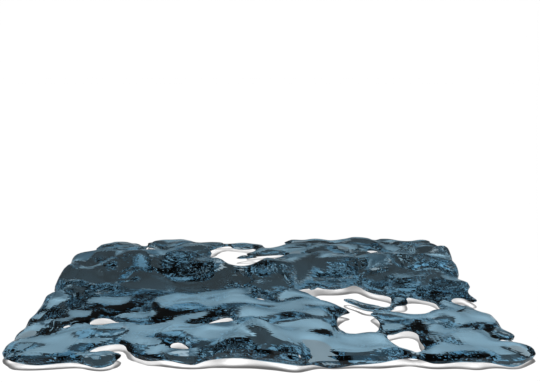}
\captionof{figure}{Frames from a simulation reconstructed with our framework. The model is able to capture the drastic topological changes in the deformed geometry.}
\label{fig:fluid-topology-changes}

%% file: imgs/fluid-dfaust-examples/figure.tex
\centering
\includegraphics[width=0.33\columnwidth]{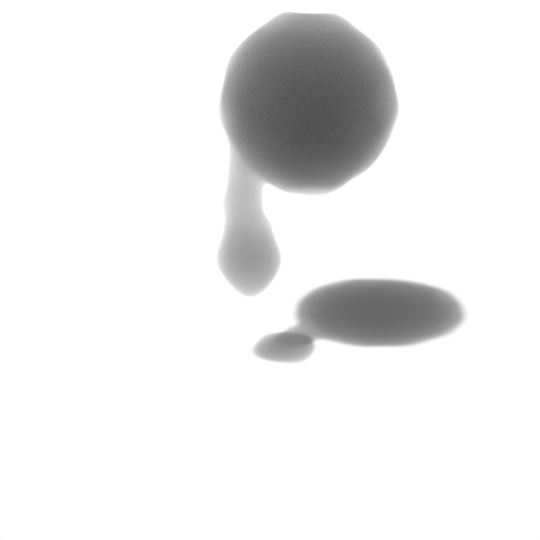}%
\includegraphics[width=0.33\columnwidth]{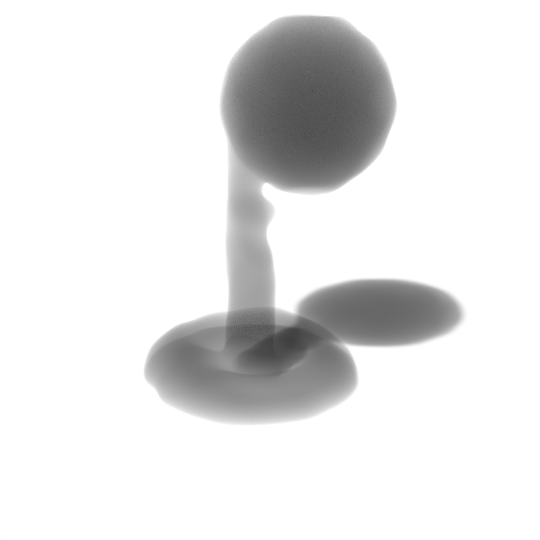}%
\includegraphics[width=0.33\columnwidth]{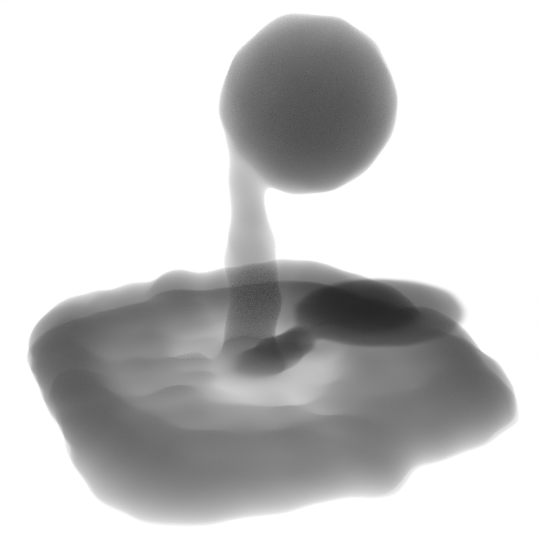}
\includegraphics[width=0.32\columnwidth]{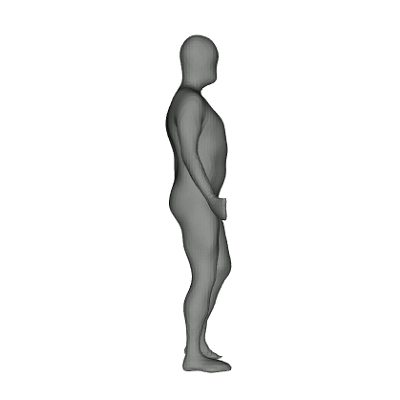}
\includegraphics[width=0.32\columnwidth]{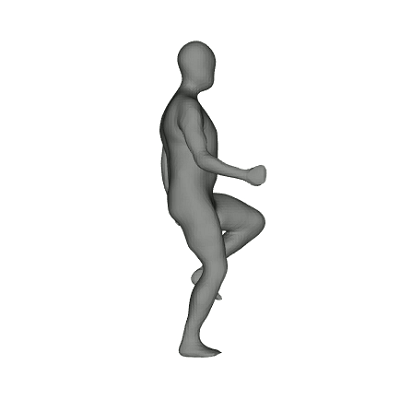}
\includegraphics[width=0.32\columnwidth]{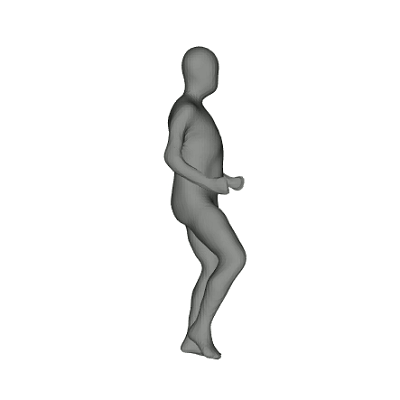}
\captionof{figure}{Our neural implicit representation can infer the dynamics of complex fluid simulation scenes (left) and simpler human body animations (right) with a unified formulation, while defining correspondences among the reconstructed surfaces.}
\label{fig:scenes-geometry-vecfield}

%% file: imgs/cold-gas-obstacles/figure.tex
\centering
\includegraphics[width=.33\columnwidth]{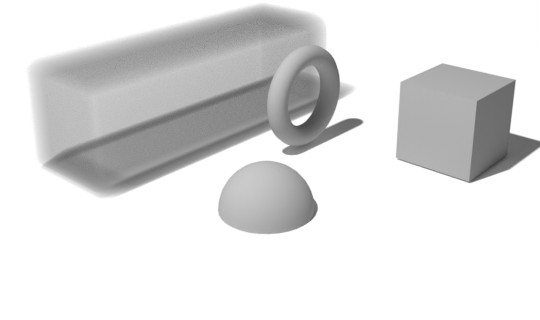}%
\includegraphics[width=.33\columnwidth]{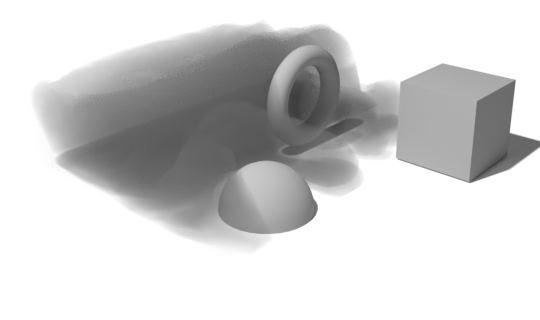}%
\includegraphics[width=.33\columnwidth]{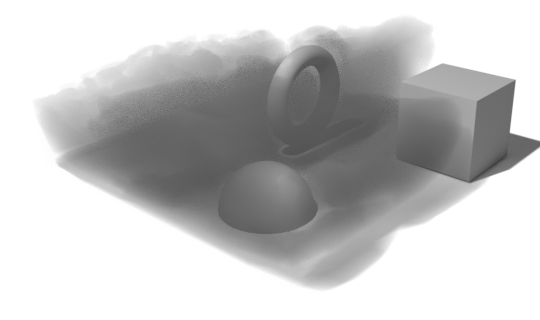}
\includegraphics[width=.33\columnwidth]{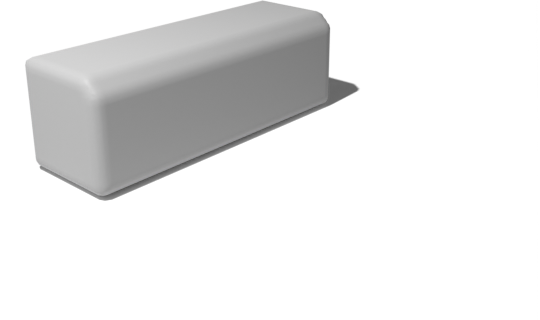}%
\includegraphics[width=.33\columnwidth]{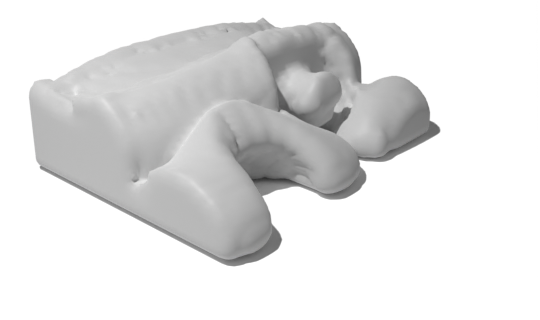}%
\includegraphics[width=.33\columnwidth]{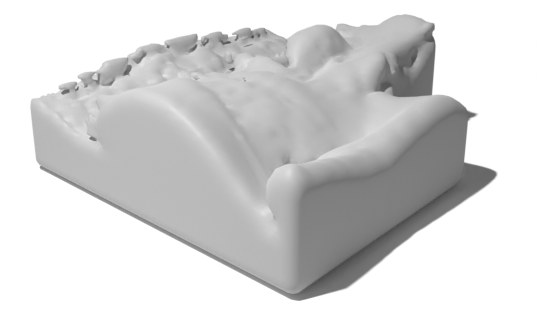}
\includegraphics[width=.33\columnwidth]{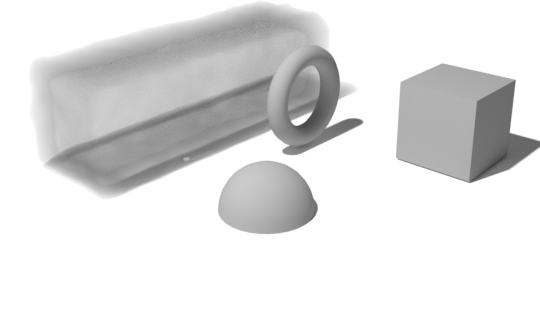}%
\includegraphics[width=.33\columnwidth]{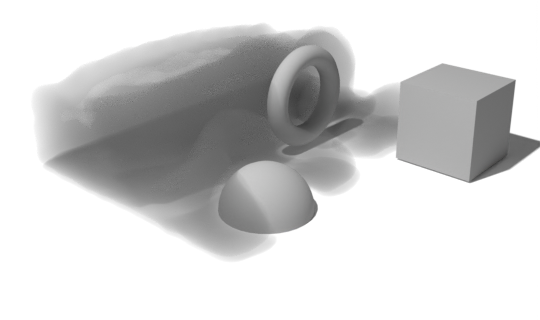}%
\includegraphics[width=.33\columnwidth]{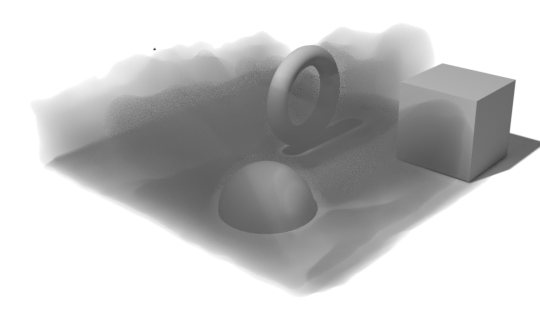}
\includegraphics[width=.33\columnwidth]{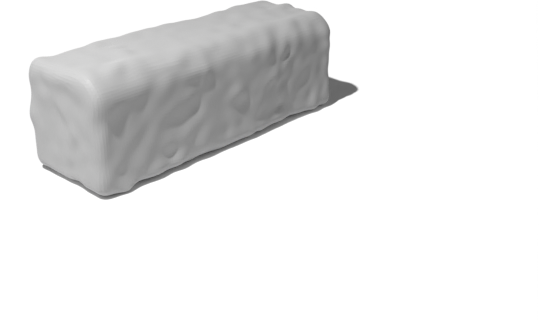}%
\includegraphics[width=.33\columnwidth]{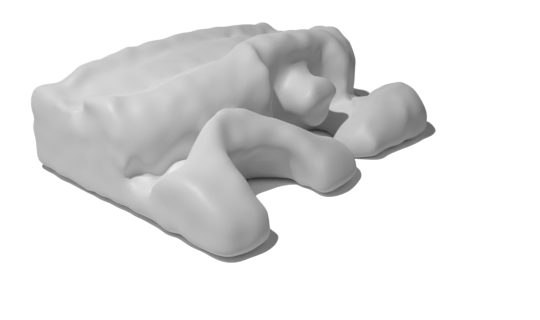}%
\includegraphics[width=.33\columnwidth]{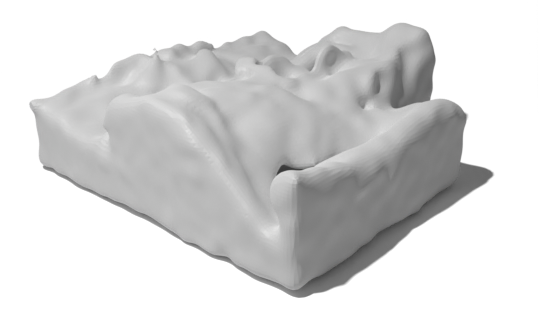}
\caption{The simulation of a cold gas in a scene containing obstacles (1st row) and the geometry surrounding the volume (2nd row), compared with the reconstruction obtained with our framework (3rd row) and the reconstructed mesh (4th row). Our method is able to capture the complex behaviour of the gas moving around the obstacles in the scene.}
\label{fig:obstacles-reconstruction}

%% file: imgs/dfaust-reconstruction/figure.tex
\centering
\includegraphics[width=\columnwidth, trim=10 120 10 120, clip]{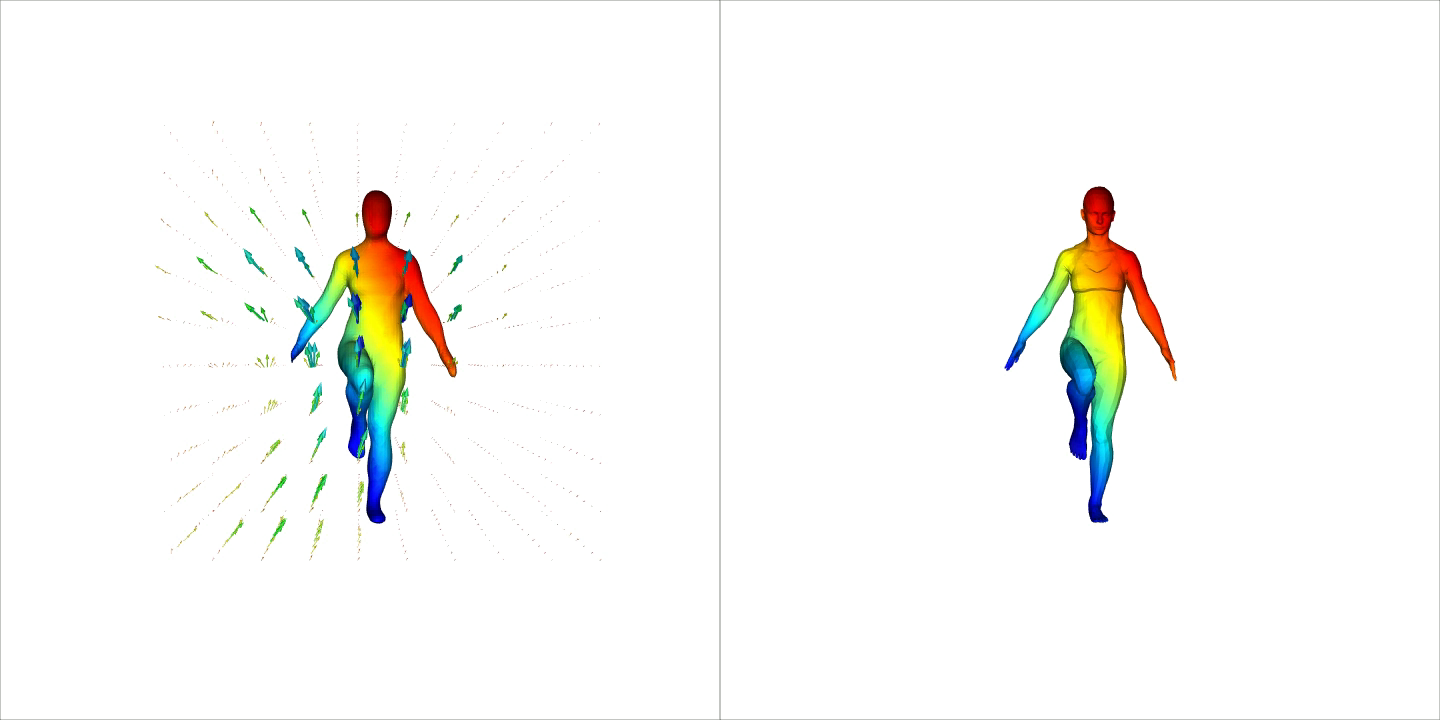}
\includegraphics[width=\columnwidth, trim=10 120 10 120, clip]{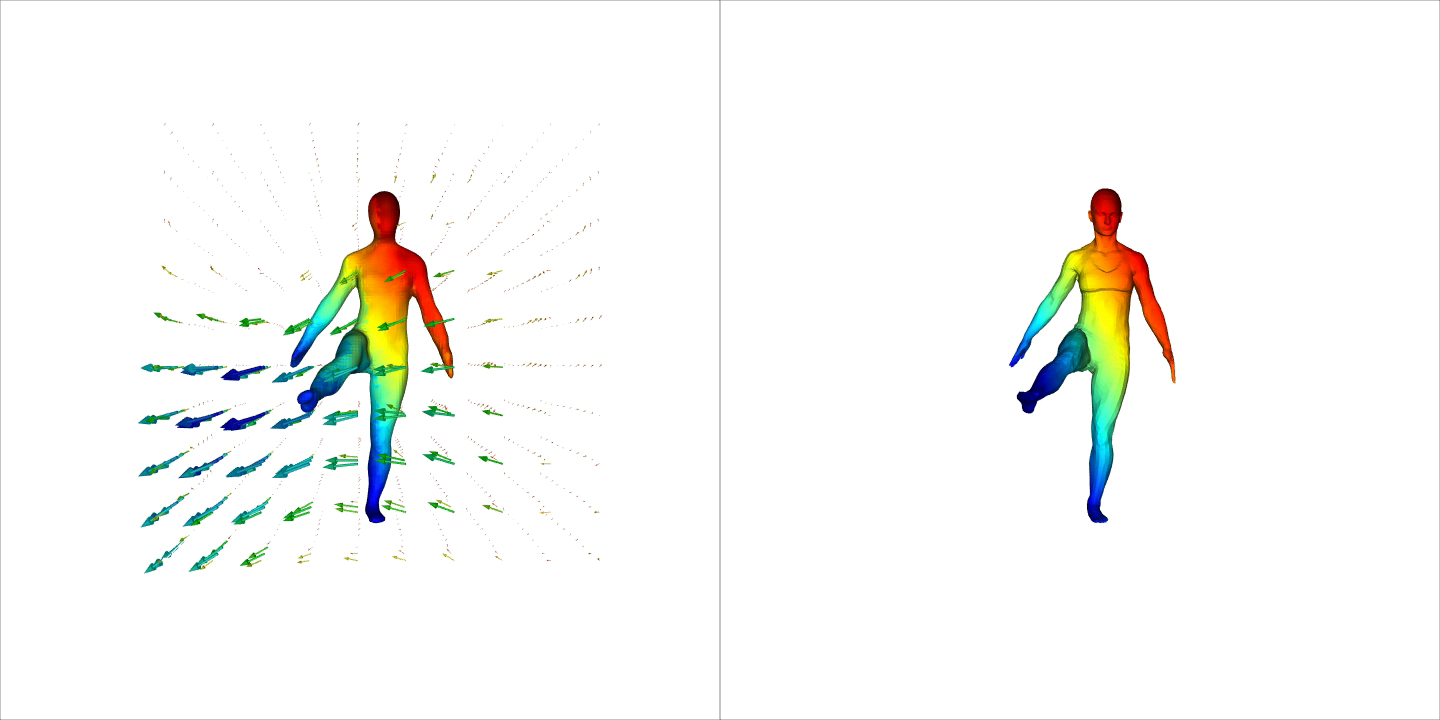}
\includegraphics[width=\columnwidth, trim=10 120 10 120, clip]{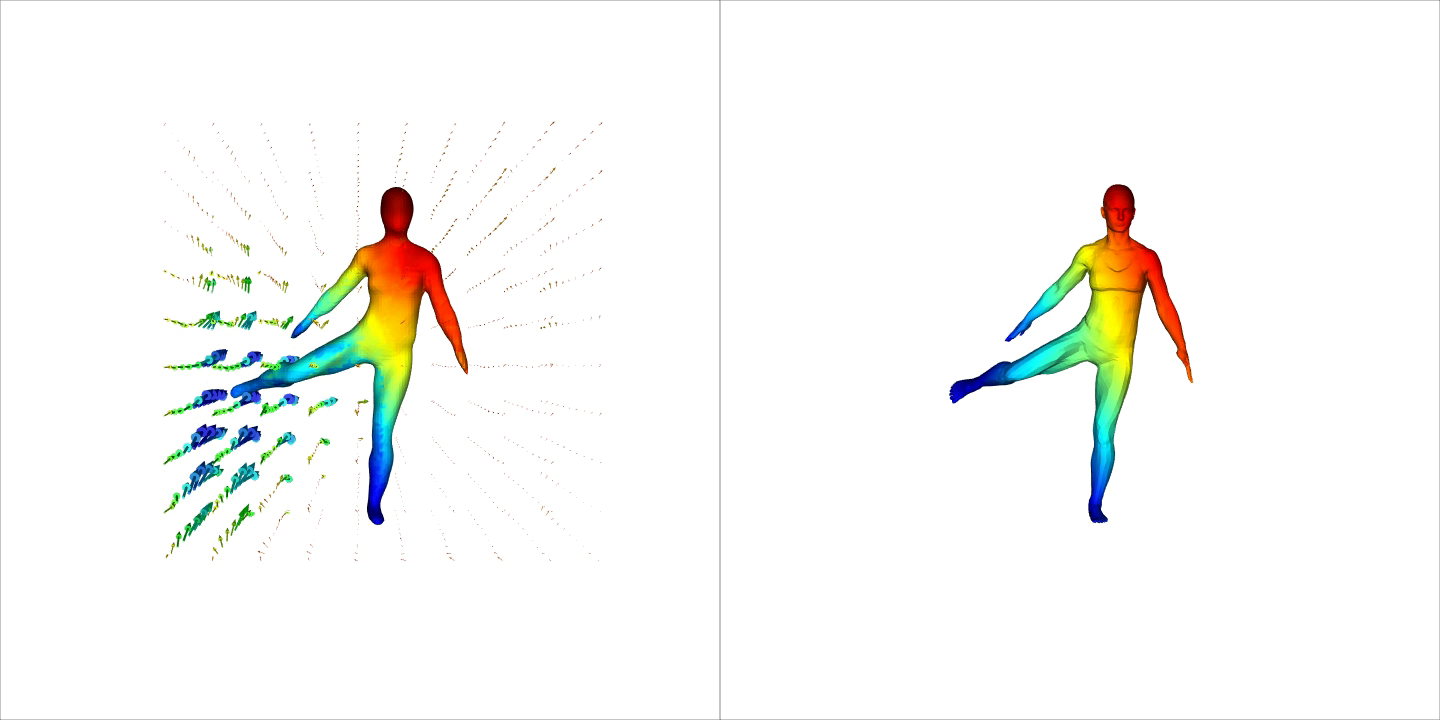}
\caption{An animation from the D-FAUST dataset (right) compared with the reconstruction obtained with our model (left). The colormap shows the preservation of correspondences established by the velocity function along the deformation.}
\label{fig:dfaust-reconstruction}

%% file: imgs/viscous-vortex/figure.tex
\centering
\includegraphics[width=.33\columnwidth]{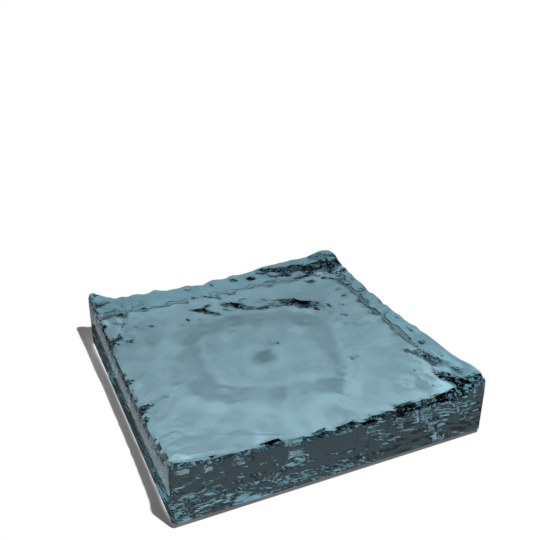}%
\includegraphics[width=.33\columnwidth]{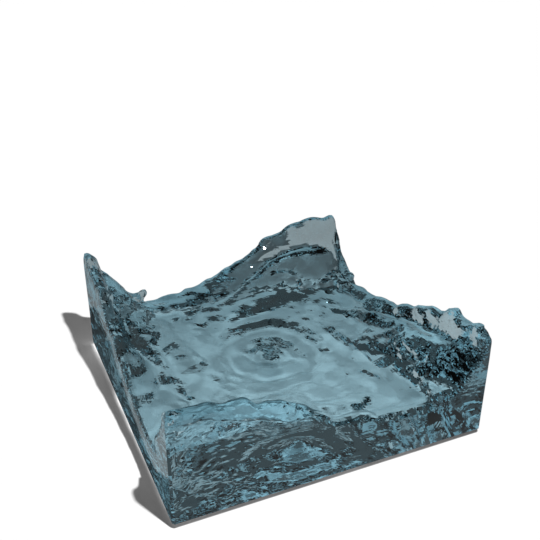}%
\includegraphics[width=.33\columnwidth]{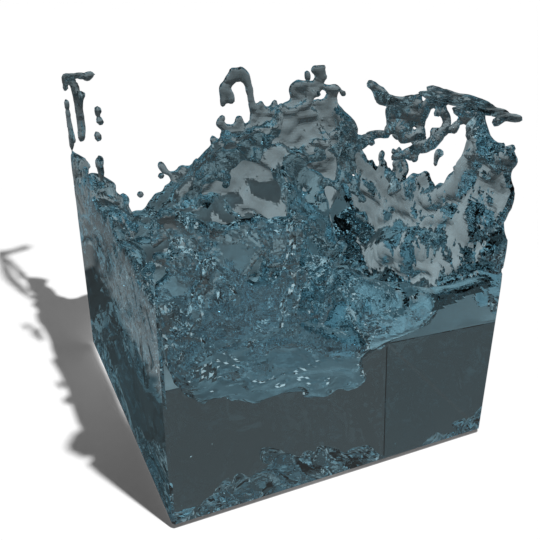}
\includegraphics[width=.33\columnwidth]{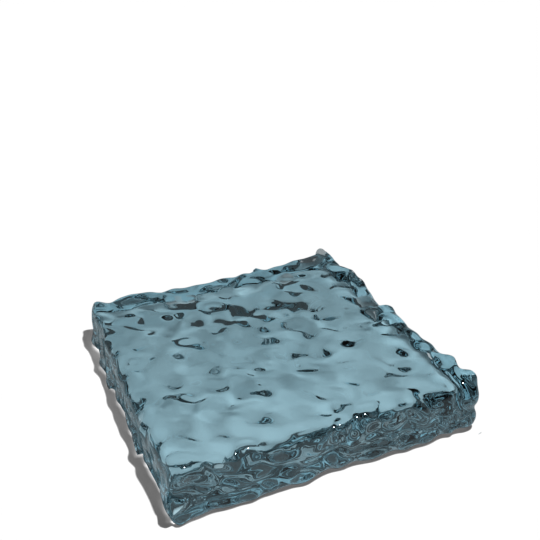}%
\includegraphics[width=.33\columnwidth]{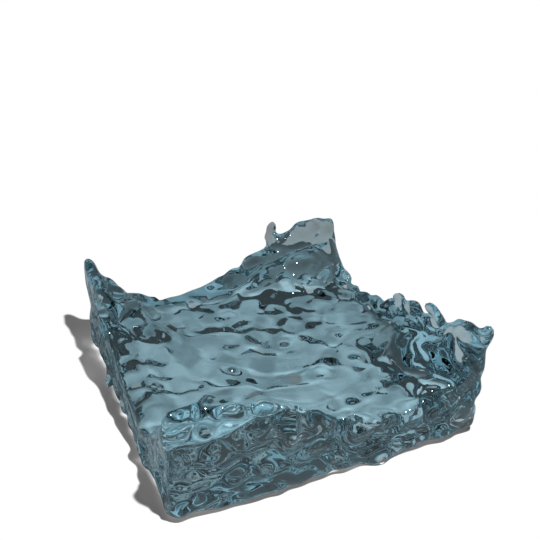}%
\includegraphics[width=.33\columnwidth]{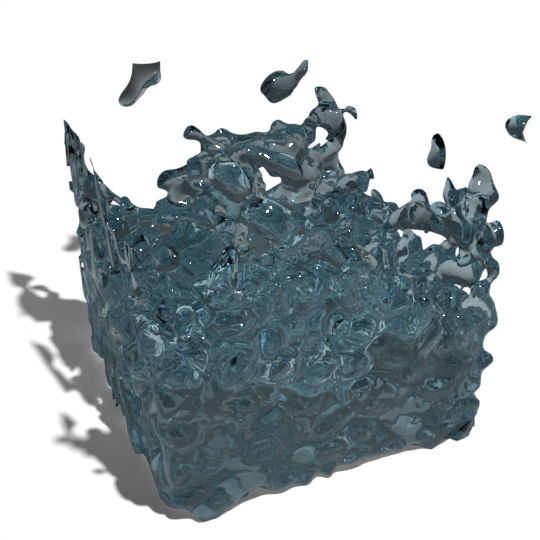}
\caption{Real simulation of a swirling fluid (top) compared with the reconstruction obtained with our method (bottom). The fluid dynamics priors given to the model allow for catching the complexity in the deformation.}
\label{fig:fluid-sim-train-compare}

%% file: imgs/ball-siren-compare/figure.tex
\centering
\begin{overpic}[width=0.33\columnwidth]{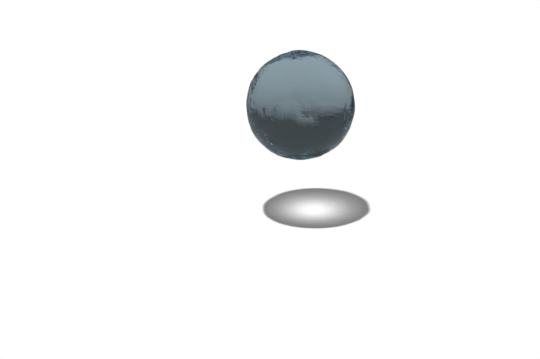}
\put(0, 35) {$\substack{\text{{\normalsize ground}}\\\text{{\normalsize truth}}}$}
\end{overpic}%
\includegraphics[width=.33\columnwidth]{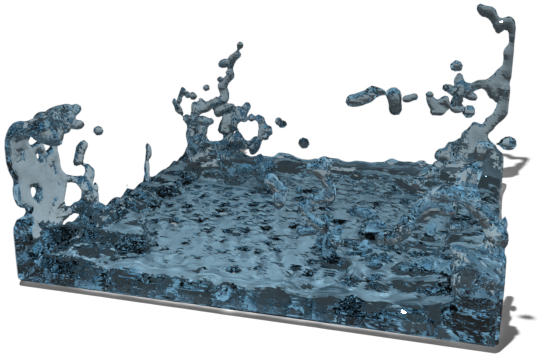}%
\includegraphics[width=.33\columnwidth]{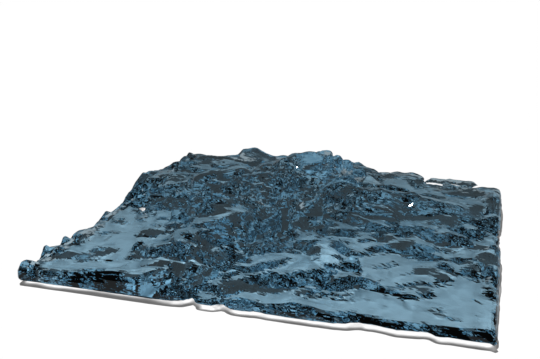}
\begin{overpic}[width=0.33\columnwidth]{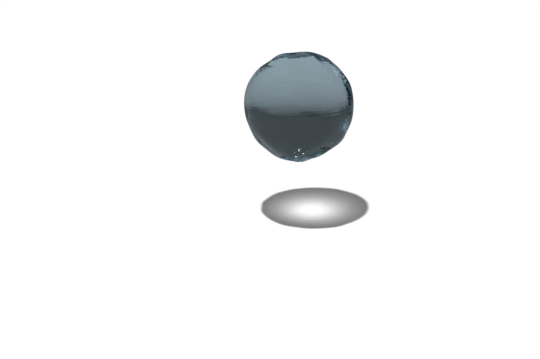}
\put(4, 35) {Siren}
\end{overpic}%
\includegraphics[width=.33\columnwidth]{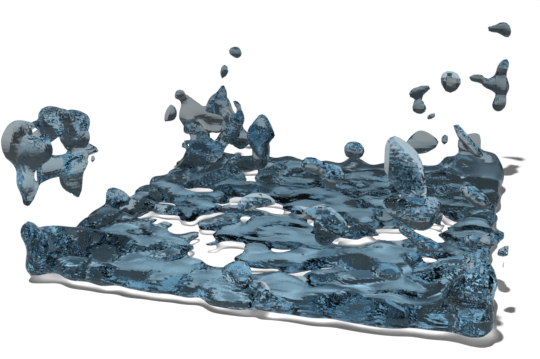}%
\includegraphics[width=.33\columnwidth]{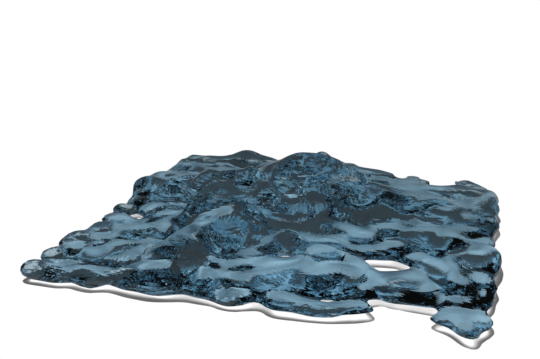}
\begin{overpic}[width=0.33\columnwidth]{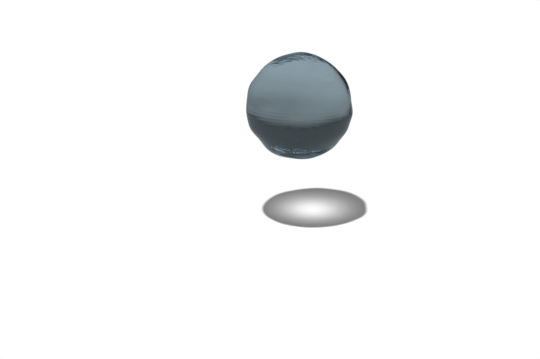}
\put(6, 35) {MLP}
\end{overpic}%
\includegraphics[width=.33\columnwidth]{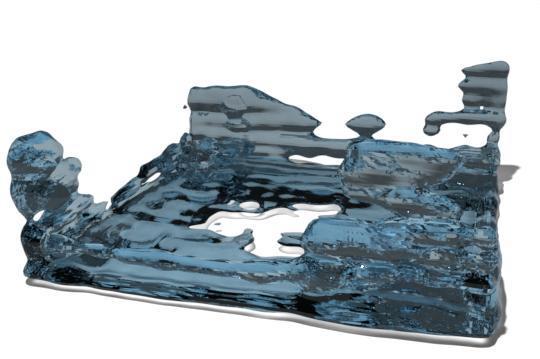}%
\includegraphics[width=.33\columnwidth]{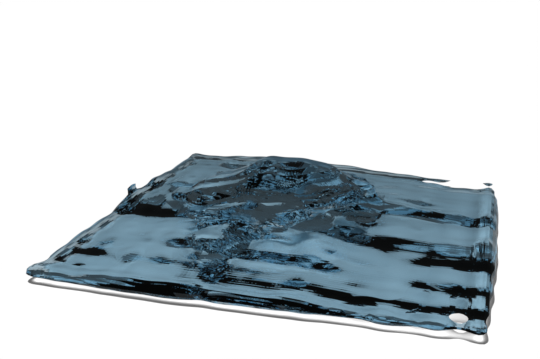}
\caption{Frames from a simulation obtained using the Siren networks (middle) compared with Softplus-activated MLPs (bottom). The former is able capture higher frequency details, producing surface displacements typical of moving fluids, and better matching the reference simulation (top).}
\label{fig:fluid-siren-vs-prev}

%% file: imgs/non-homeo-matching/figure.tex
\centering
\includegraphics[width=\columnwidth, trim=100 175 100 200, clip]{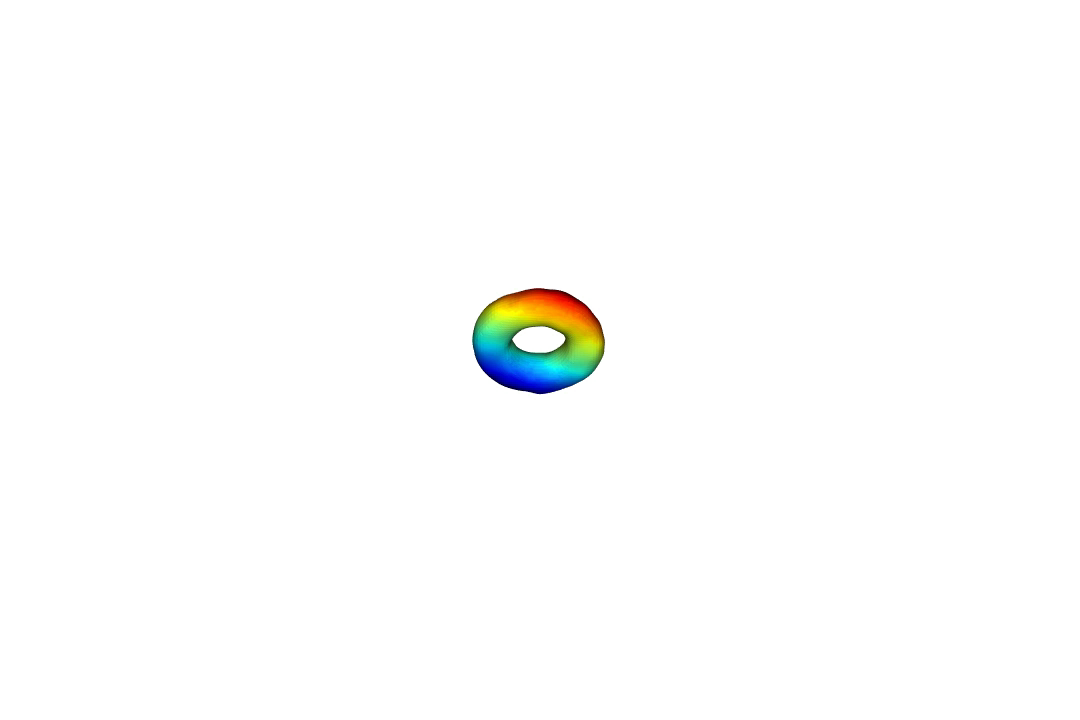}
\includegraphics[width=\columnwidth, trim=100 50 100 200, clip]{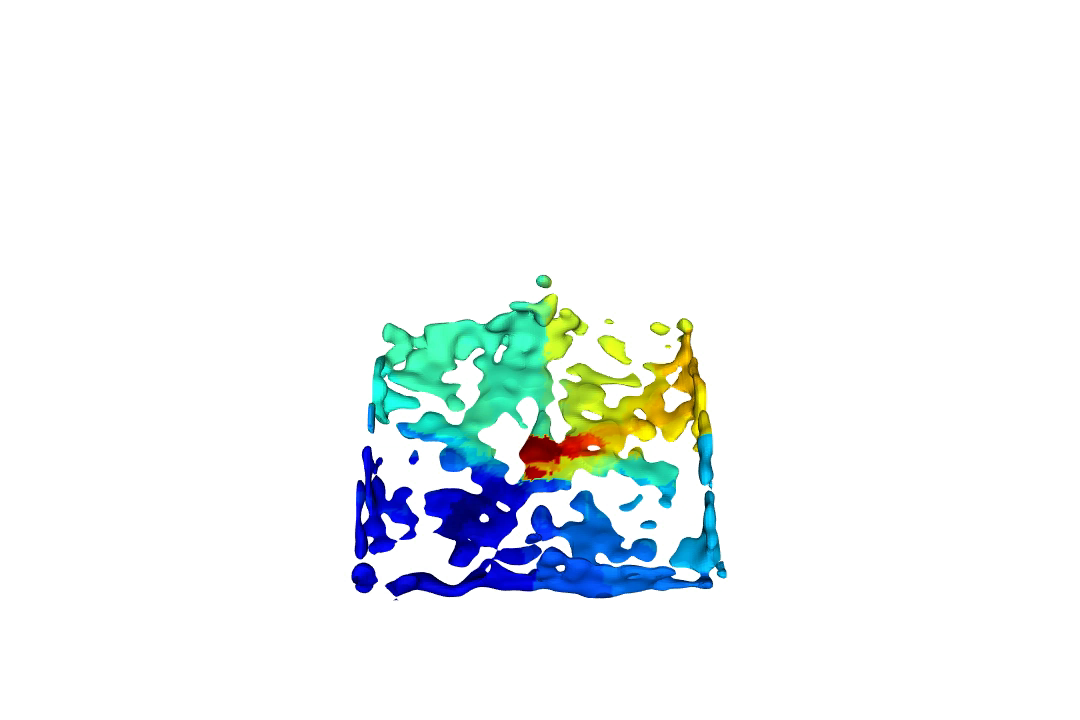}
\includegraphics[width=\columnwidth, trim=100 50 100 200, clip]{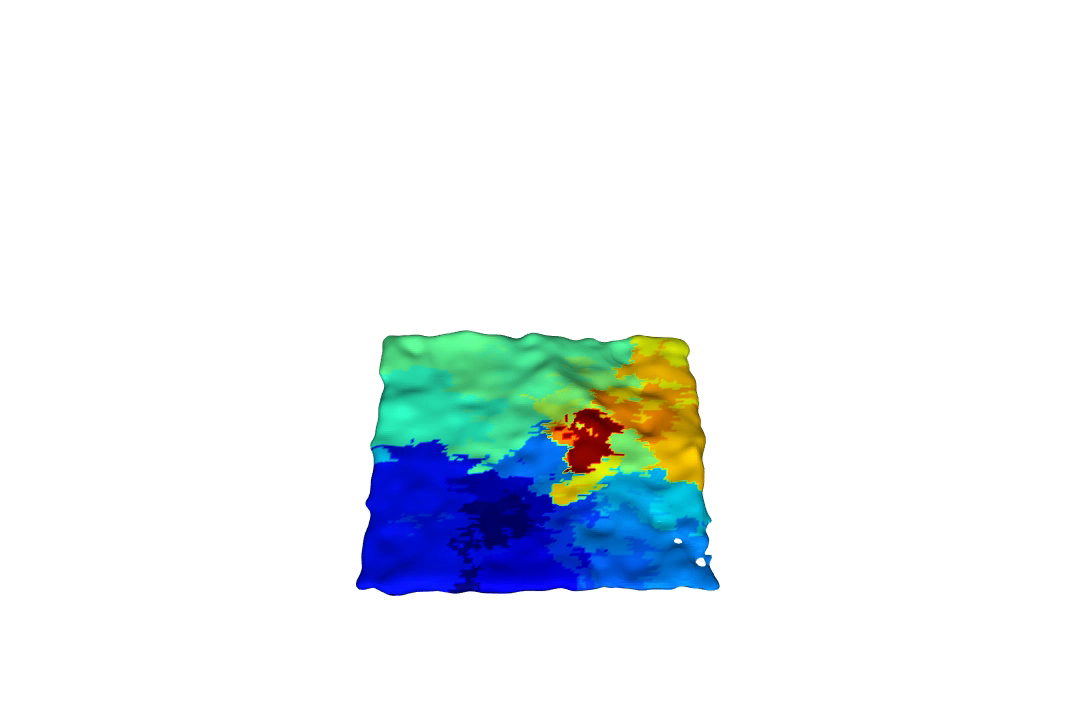}
\caption{An example of predicted correspondences under dramatic changes of topology. Despite the absence of semantics in vertex identity, the indicator function is preserved coherently with surface dynamics.}
\label{fig:non-homeo-matching}